\documentclass[letterpaper]{article} 
\usepackage[preprint]{aaai2027}  
\usepackage[hyphens]{url}  
\usepackage{graphicx} 
\usepackage{natbib}  
\usepackage{caption} 
\usepackage{algorithm}
\usepackage{algorithmic}
\usepackage{amssymb}
\usepackage{amsmath}
\usepackage{multirow}
\usepackage{booktabs}
\usepackage{makecell}
\usepackage{newfloat}
\usepackage{listings}
\DeclareCaptionStyle{ruled}{labelfont=normalfont,labelsep=colon,strut=off} 
\floatstyle{ruled}
\newfloat{listing}{tb}{lst}{}
\floatname{listing}{Listing}

\usepackage{booktabs}
\usepackage[table]{xcolor}

\title{Look Where It Matters: Adaptive Visual Refinement for Vision-Language-Action Models}
\author {
    Jin Cui\textsuperscript{\rm 1}\equalcontrib,
    Yanbin Hu\textsuperscript{\rm 1, \rm 2}\equalcontrib,
    Xinyue Long\textsuperscript{\rm 1, \rm 2},
    Linkai Li\textsuperscript{\rm 1, \rm 2},
    Boran Zhao\textsuperscript{\rm 1,\rm 2}\corresponding,
    Pengju Ren\textsuperscript{\rm 1}
}
\affiliations {
    \textsuperscript{\rm 1}State Key Laboratory of Human-Machine Hybrid Augmented Intelligence,\\
    and Institute of Artificial Intelligence and Robotics, Xi'an Jiaotong University\\
    \textsuperscript{\rm 2}School of Software Engineering, Xi'an Jiaotong University \\
    andycui@stu.xjtu.edu.cn
}

\begin{document}

\maketitle

\begin{abstract}
Vision-language-action (VLA) models inherit rich semantic priors from large-scale vision-language pretraining, yet their visual representations remain unreliable for spatially precise robotic manipulation. We uncover that vision encoders in VLAs also exhibit attention artifacts previously documented in generic Vision Transformers, and further show that, in embodied policies, these artifacts are closely associated with spatial perception capabilities acquired during post-training. As the encoder learns task-relevant information such as object location, depth ordering, and local geometry, limited global-token capacity causes part of this information to spill into low-information patch tokens. We introduce \textit{\textbf{AtVLA}}, a framework that inserts learnable register tokens into the visual encoder. Trained end-to-end using only embodied data and the original action objective, these registers emerge as dedicated carriers of embodied spatial information, while the remaining patch tokens recover clean and spatially faithful attention distributions crucial for precise target localization and fine-grained contact. Clean attention restores reliable localization, but cannot recover geometric details lost in low-resolution observations. \textit{\textbf{AtVLA}} therefore couples attention rectification with uncertainty-gated local refinement. The action expert samples multiple action chunks and estimates uncertainty from their disagreement; only for uncertain predictions, action-conditioned attention rollout identifies the task-relevant region, which is cropped, re-encoded at high resolution, and appended to the cached prefix for refined action generation. Across LIBERO, SimplerEnv, and a challenging single-view real-world benchmark, AtVLA improves the average LIBERO success rate from 94.2\% to 98.4\% and real-world success from 46.5\% to 69.0\% over $\pi_0$. The additional inference cost remains limited: cropping is triggered on approximately 30\% of replanning steps, resulting in only $1.4$--$1.6\times$ the total computation of $\pi_0$ under the representative deployment setting.
\end{abstract}


\section{Introduction}
Vision-language-action (VLA) models have emerged as a promising paradigm for generalist robot manipulation, transferring the semantic knowledge and instruction-following capabilities of large-scale vision-language models into continuous control policies \cite{Kim2024OpenVLAAO, brohan2023rt2visionlanguageactionmodelstransfer, Black20240AV}. By post-training pretrained vision-language backbones on diverse robot demonstrations, recent VLAs achieve strong generalization across manipulation tasks. However, robotic control requires more than semantic recognition; the visual encoder must preserve spatially faithful representations for target localization, geometric reasoning, and precise contact. This raises a fundamental question: \emph{how does embodied post-training reshape the visual representations of pretrained encoders, and are these representations sufficiently reliable for fine-grained manipulation?}


Through systematic analysis of manipulation failures, we identify two visual bottlenecks in current VLAs. \textbf{\textit{First, models frequently fail to correctly identify target objects and disambiguate task types}}---a surprising deficiency given the powerful semantic understanding of their VLM backbones. Inspecting their attention distributions, we find that VLA visual encoders exhibit the high-norm artifacts previously observed in generic Vision Transformers \cite{Darcet2023VisionTN}: background patches are repurposed as repositories for global information, displacing local spatial content and corrupting dense attention maps. We further show that this pathology is closely associated with embodied post-training. As the encoder acquires manipulation-relevant information such as object location, depth ordering, instance structure, and local geometry, the limited capacity of its original global tokens causes part of this information to spill into spatial patch tokens. Useful embodied knowledge is therefore stored in the wrong representational channels, undermining the precise localization and contact reasoning required for control.

\textbf{\textit{Second, even correct localization does not guarantee precise manipulation.}} Fine-grained actions require local geometric details near the target and contact region, yet such details may occupy only a few patches in a low-resolution third-person observation. Wrist-mounted cameras provide close-range views, but suffer from acute viewpoint and pose-change sensitivity; the highly variable relative position between the gripper and object causes the relevant image region to shift unpredictably. Explicit vision experts or 3D representations can provide stronger geometry, but introduce additional sensing, calibration, and modeling complexity. These observations motivate a selective coarse-to-fine mechanism that first restores reliable localization and then increases visual resolution only when the base observation is insufficient.

We introduce \textit{\textbf{AtVLA}}, a framework that combines register-enhanced visual encoding with uncertainty-gated local refinement. We insert learnable register tokens into the visual encoder and optimize them using only embodied demonstrations and the original action objective. During post-training, the registers emerge as dedicated carriers of spatial information acquired from robot interaction, while the remaining patch tokens recover clean and spatially faithful attention distributions. Unlike discarding these tokens, we retain their outputs as additional spatial context for action generation. Linear-probing confirms that they encode stronger task-relevant spatial information than the original class token or pooled patch features. Based on the rectified visual prefix, the action expert samples multiple action chunks and estimates uncertainty from their disagreement. Confident predictions follow the original inference path. For uncertain predictions, \textit{\textbf{AtVLA}} reuses action-conditioned attention to localize the task-relevant region, crops and re-encodes it at high resolution, and appends the resulting tokens to the cached prefix for refined action generation. Clean attention determines \emph{where} to look, while adaptive cropping recovers the local geometry needed to act precisely.

\begin{figure}
    \centering
    \includegraphics[width=1\linewidth]{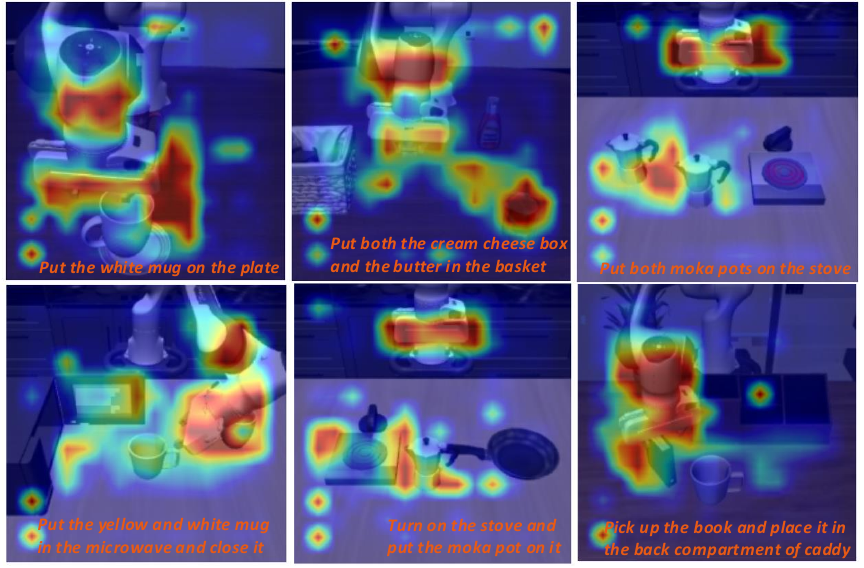}
    \caption{Attention artifacts in a standard VLA visual encoder. High-norm regions emerge in low-information backgrounds or unrelated objects, corrupting the spatial grounding of task-relevant targets and interaction regions.}
    \label{fig:artifacts}
\end{figure}

Across LIBERO, SimplerEnv, and a challenging real-world benchmark, \textit{\textbf{AtVLA}} consistently improves object grounding, spatial reasoning, and fine-grained manipulation. It raises the average LIBERO success rate of $\pi_0$ from $94.2\%$ to $98.4\%$ and real-world success from $46.5\%$ to $69.0\%$. Local refinement is triggered on only approximately $30\%$ of replanning steps, yielding $1.4$--$1.6\times$ the total computation of $\pi_0$ under our deployment setting. Our contributions are summarized as follows:
\begin{itemize}
     \item We uncover attention artifacts in VLA visual encoders and reveal their connection to spatial features acquired during embodied post-training. After rectification, registers emerge as stable carriers of embodied spatial information, while restoring clean, spatially faithful attention.
    \item We develop uncertainty-gated, attention-guided visual refinement that selectively crops and re-encodes task-relevant regions, enabling high-resolution perception only when needed for precise manipulation.
    \item Extensive experiments across simulated and real-world benchmarks demonstrate substantial performance gains with modest overhead, as adaptive refinement is invoked on only a fraction of replanning steps.
\end{itemize}

\section{Related Work}

\paragraph{Attention Artifacts in Visual Encoders and VLA Grounding.} Patch-level representations are critical for spatial grounding in vision-language-action models. Large-scale Vision Transformers can develop high-norm tokens in low-information regions, which act as repositories for global information and distort dense spatial features \cite{Darcet2023VisionTN}. We find that this phenomenon is further exacerbated during embodied post-training, where visual encoders must capture task-relevant geometry, instance structure, and spatial relations beyond conventional 2D semantics \cite{Qu2025SpatialVLAES,pmlr-v305-li25g,Li2025PointVLAIT}. Related studies report attention sinks, representation collapse, and dispersed task attention in VLA policies \cite{Kachaev2025DontBY,Song2025ReconVLARV}, while analyses of VLMs show that visual grounding signals can deteriorate across network depth \cite{Kaduri2024WhatsIT,Jiang2024DevilsIM}. Object-centric tokenization and slot-based representations improve VLA grounding through feature selection or decomposition \cite{Bendikas2025FocusingOW,Hanyu2025SlotVLATM}, but do not explicitly address corrupted patch representations. In contrast, we use register tokens to absorb excess global information while preserving clean patch features for spatial grounding.

\begin{figure*}
    \centering
    \includegraphics[width=1\linewidth]{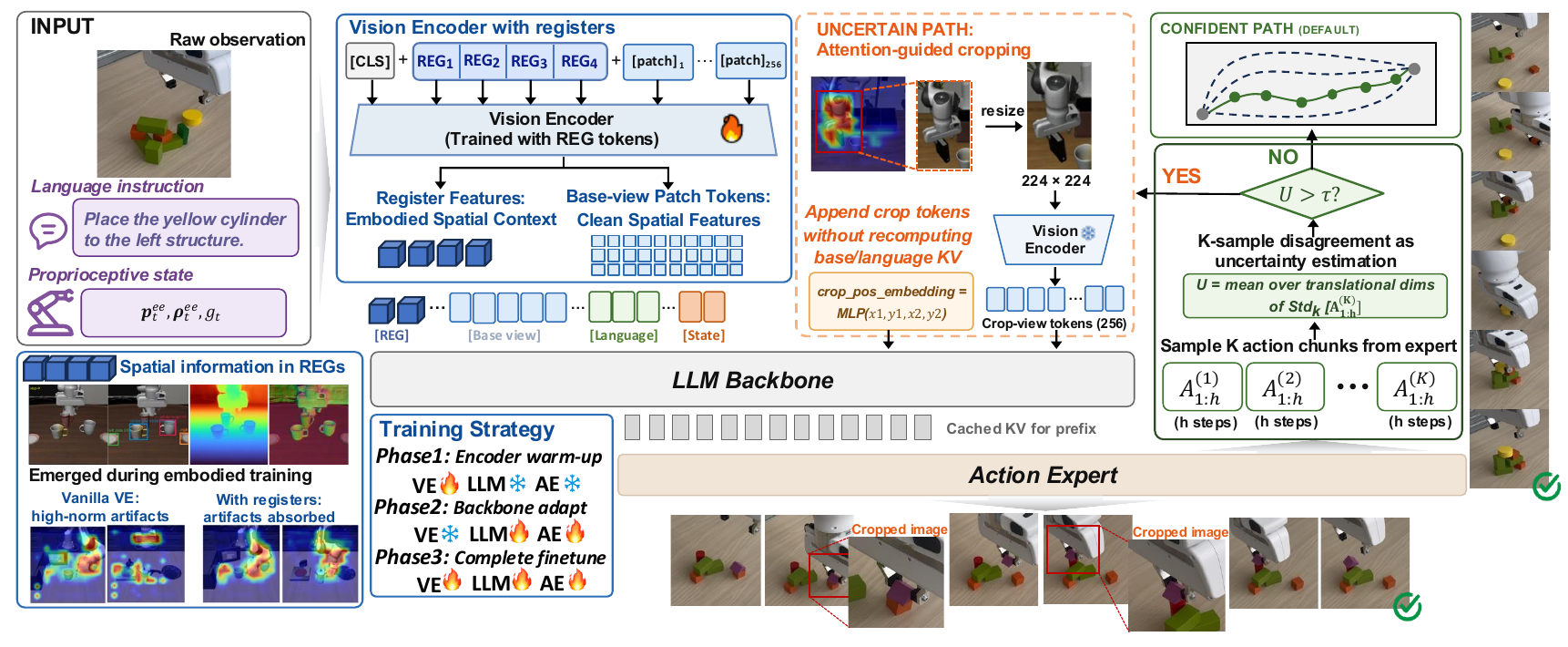}
    \caption{Overview of \textit{\textbf{AtVLA}}. Register tokens absorb excess global information and retain embodied spatial context while restoring clean patch representations. The action expert estimates uncertainty from multiple sampled action chunks: confident predictions follow the base path, whereas uncertain predictions trigger attention-guided cropping, high-resolution re-encoding, and prefix extension. A three-stage curriculum progressively aligns the registers, crop branch, and complete policy.}
    \label{fig:placeholder}
\end{figure*}

\paragraph{Visual Refinement for Precise Perception.}
MLLMs may correctly localize task-relevant regions yet fail to resolve fine details under limited input resolution, motivating targeted visual re-examination \cite{Zhang2025MLLMsKW}. Previous work explores hierarchical search, attention-guided cropping, KV-cache-based localization, and learned zoom policies \cite{Shen2024ZoomEyeEM,Liu2024ChainofSpotIR,zhang2024perceivingsmallvisualdetails,zhong2025focusinternalmllmrepresentations,carvalho2026cropvlmlearningzoomfinegrained}. In robotic manipulation, related methods incorporate auxiliary visual cues or explicit spatial supervision to improve manipulation grounding \cite{Li2024VIPVI,Zawalski2024RoboticCV,Deng2025GraspVLAAG}, but do not directly leverage the policy's internal attention for adaptive visual refinement. We instead use rectified attention maps for uncertainty-gated cropping, re-encoding task-relevant regions at high resolution only when the base action prediction is uncertain.

\section{Method}
\label{sec:method}

\subsection{Overview}

Given a raw visual observation $\mathbf{I}_t$, a language instruction $\ell$, and a proprioceptive state $\mathbf{s}_t$, our method augments $\pi_0$ with two complementary mechanisms: register-enhanced visual encoding and uncertainty-gated local visual refinement. Register tokens
provide dedicated global-information slots inside the SigLIP encoder, preventing background patch tokens from being repurposed as high-norm information reservoirs. Their output features are retained as multi-slot global context, while the remaining patch tokens preserve spatially faithful local representations. Based on this clean visual prefix, the
flow-matching action expert first predicts multiple action chunks and estimates action uncertainty from their disagreement. The base prediction is executed directly when uncertainty is low. Otherwise, an attention rollout localizes the task-relevant region, which is re-encoded at full
resolution and appended to the cached prefix for refined action generation. The gate is evaluated once per action-chunk replanning step rather than at every low-level control step.

\subsection{Register-Enhanced Visual Prefix}

Let the vision encoder operate on $224\times224$ images with patch size $14$, producing $N_b=16\times16=256$ patch tokens. We insert $N_r=4$ learnable register embeddings immediately after the class token:
\begin{equation}
    \mathbf{X}_t =
    [\mathbf{x}_{\mathrm{cls}};
     \mathbf{r}_1,\ldots,\mathbf{r}_{N_r};
     \mathbf{p}_1,\ldots,\mathbf{p}_{N_b}].
\end{equation}
The register-augmented visual encoder $E_{\mathrm{vis}}$ outputs
\begin{equation}
    [\mathbf{h}_{\mathrm{cls}},\mathbf{R}_t,\mathbf{P}_t]
    = E_{\mathrm{vis}}(\mathbf{X}_t),
\end{equation}
where $\mathbf{R}_t\in\mathbb{R}^{N_r\times d_v}$ denotes register
features and $\mathbf{P}_t\in\mathbb{R}^{N_b\times d_v}$ denotes
spatial patch features. Unlike the original use of registers solely as
internal workspace, we retain $\mathbf{R}_t$ and project it through the
PaliGemma visual projector $\Pi$. The base multimodal prefix is
\begin{equation}
    \mathbf{C}_t^{b} =
    [\Pi(\mathbf{R}_t);
     \Pi(\mathbf{P}_t);
     E_{\mathrm{lang}}(\ell);
     E_{\mathrm{state}}(\mathbf{s}_t)].
    \label{eq:base-prefix}
\end{equation}
Thus, the registers expose several global context slots to the downstream
policy, while the full $16\times16$ patch grid remains available for
spatial grounding. For multiple camera views, each view is encoded
independently with the same register-augmented encoder, and the resulting
visual sequences are concatenated in the prefix.

\subsection{Uncertainty-Gated Action Prediction}

LLM backbone processes $\mathbf{C}_t^{b}$ once to construct a reusable
prefix key--value cache. Conditioned on this cache, the flow-matching
action expert generates $K$ action chunks from independent Gaussian
initializations:
\begin{equation}
    \mathbf{z}^{(k)}_0\sim\mathcal{N}(\mathbf{0},\mathbf{I}),\qquad
    \mathbf{A}^{(k)}_{1:H}
    = \operatorname{ODESolve}
    \bigl(v_{\theta},\mathbf{z}^{(k)}_0\mid\mathbf{C}_t^{b}\bigr)
\end{equation}
where $H$ is the predicted chunk horizon. All samples share the visual and language prefill and differ only in action-expert denoising. We measure disagreement over the translational action dimensions
$\mathcal{D}_{\mathrm{tr}}$ and the first $h\leq H$ actions that will
actually be executed:
\begin{equation}     \small
    U_t
    = \frac{1}{h|\mathcal{D}_{\mathrm{tr}}|}
       \sum_{j=1}^{h}\sum_{d\in\mathcal{D}_{\mathrm{tr}}}
       \sqrt{\frac{1}{K-1}
       \sum_{k=1}^{K}
       \left(A^{(k)}_{j,d}-\frac{1}{K}\sum_{k=1}^{K} A^{(k)}_{j,d}\right)^2}
    \label{eq:uncertainty}
\end{equation}
Restricting the score to near-term translation isolates disagreement in the end-effector motion most directly related to spatial ambiguity. The noise schedule, ODE solver, and number of integration steps are fixed when computing $U_t$. If $U_t\leq\tau$, the mean action chunk $\overline{\mathbf{A}}_{1:H}$ is returned through the default confident path. The threshold $\tau$ is selected on a held-out set by measuring how well $U_t$ predicts the empirical error reduction obtained from cropping.

\begin{table*}[t]
  \caption{
    Success rates (\%) on commonly used benchmarks with standard deviations reported when available.
    Baseline results are taken from reported settings.
    PCC, MN and O/C denote Pick Coke Can, Move Near and Open/Close. \textit{$\pi_0$ + Cropping} variant takes cropped images from an external VLM without embodied pre-training, while $^*$ is analyzed in the following context.
}
  \vspace{-2mm}
  \label{tab:main_results}
  \centering
  \small
  \setlength{\tabcolsep}{6pt}
  \begin{tabular}{lcccclccc}
    \toprule
    \multirow{2}{*}{Method} & \multicolumn{4}{c}{Standard LIBERO Suites} & \multirow{2}{*}{Method} & \multicolumn{3}{c}{SimplerEnv Google}\\
    \cmidrule(r){2-5}
    \cmidrule(l){7-9}
    & Spatial & Object & Goal & Long-10 
    &
    & PCC & MN & O/C \\
    \midrule
    Octo~\cite{Team2024OctoAO} & 78.9$_{\pm \text{1.0}}$ & 85.7$_{\pm \text{1.2}}$ & 84.6$_{\pm \text{0.9}}$ & 50.9$_{\pm \text{1.2}}$ & Octo~\cite{Team2024OctoAO} & 17.0 & 4.2 & 22.7 \\
    
    OpenVLA~\cite{Kim2024OpenVLAAO} & 85.0$_{\pm \text{1.1}}$ & 88.6$_{\pm \text{0.9}}$ & 79.2$_{\pm \text{1.0}}$ & 53.6$_{\pm \text{1.0}}$ & OpenVLA~\cite{Kim2024OpenVLAAO} & 16.3 & 46.2 & 35.6 \\
    
    $\pi_0$~\cite{Black20240AV} & 96.8$_{\pm \text{1.3}}$ & 98.8$_{\pm \text{1.2}}$ & 95.8$_{\pm \text{1.5}}$ & 85.2$_{\pm \text{1.0}}$ & $\pi_0$~\cite{Black20240AV} & 88.0 & 80.3 & 56.0 \\
    
    $\pi_{0.5}$~\cite{Intelligence202505AV} & 97.3$_{\pm \text{1.7}}$ & 98.8$_{\pm \text{0.9}}$ & 96.9$_{\pm \text{1.2}}$ & 94.2$_{\pm \text{1.8}}$ & RT-1-X~\cite{brohan2023rt2visionlanguageactionmodelstransfer} & 56.7 & 31.7 & \textbf{59.7} \\
    
    SpatialVLA~\cite{Qu2025SpatialVLAES} & 88.2$_{\pm \text{0.5}}$ & 89.9$_{\pm \text{0.7}}$ & 78.6$_{\pm \text{0.6}}$ & 55.5$_{\pm \text{1.0}}$ & RT-2-X~\cite{brohan2023rt2visionlanguageactionmodelstransfer} & 78.7 & 77.9 & 25.0 \\
    
    CoT-VLA~\cite{Zhao2025CoTVLAVC} & 81.13$_{\pm \text{0.6}}$ & 87.5$_{\pm \text{1.4}}$ & 91.6$_{\pm \text{0.5}}$ & 69.0$_{\pm \text{0.8}}$ & RoboVLM~\cite{Li2024WhatMI} & 77.3 & 61.7 & 43.5 \\
    
    OFT~\citep{Kim2025FineTuningVM} & 96.2 & 98.3 & 96.2 & 90.7 & SpatialVLA~\cite{Qu2025SpatialVLAES} & 86.0 & 77.9 & 57.4 \\
    
    VLANeXt~\cite{Wu2026VLANeXtRF} & 99.0 & 99.2 & 96.6 & 94.8 & TraceVLA~\cite{Zheng2024TraceVLAVT} & 28.0 & 53.7 & 57.0 \\
    \midrule
    \rowcolor{gray!15} $\pi_0$ + Registers & 98.8$_{\pm \text{0.6}}$ & 99.0$_{\pm \text{0.8}}$ & 98.0$_{\pm \text{0.9}}$ & 93.1$_{\pm \text{1.7}}$ & $\pi_0$ + Registers & 88.2 & 80.5 & 56.0\\

    \rowcolor{gray!15} $\pi_0$ + Cropping & 96.5$_{\pm \text{1.3}}$ & 98.2$_{\pm \text{1.2}}$ & 96.7$_{\pm \text{1.5}}$ & 93.3$_{\pm \text{2.1}}$ & $\pi_0$ + Cropping & 90.0 & 79.2 & 56.9\\

    \rowcolor{gray!15} \textbf{\textit{AtVLA} (Full)} & \textbf{99.3}$_{\pm \text{0.7}}$ & \textbf{99.4}$_{\pm \text{0.8}}$ & \textbf{98.3}$_{\pm \text{1.2}}$ & \textbf{96.5}$_{\pm \text{2.6}}$ & \textbf{\textit{AtVLA} (Full)} & \textbf{91.3} & \textbf{81.6} & 57.5$^*$\\
    \bottomrule
  \end{tabular}
\end{table*}

\begin{table*}[t]

\caption{\textbf{Evaluation in RealWorld scene.} All models underwent extensive fine-tuning tailored to their respective architectures, and standard deviations were derived from 20 independent experiments.}
\vspace{-2mm}
\centering
\small
\setlength{\tabcolsep}{5pt}

\begin{tabular}{lcccccccccc}
\toprule

\multirow{2}{*}{Method}
& \multicolumn{5}{c}{Kitchen}
& \multicolumn{5}{c}{Building Blocks}
\\

\cmidrule(lr){2-6}
\cmidrule(lr){7-11}

& Move & Grab & Pick & Long & Spatial
& Stack & Edge & Spatial & Long & Grab 
\\

\midrule

Octo~\cite{Team2024OctoAO}
& 20$_{\pm \text{10}}$ & 10$_{\pm \text{5}}$ & 25$_{\pm \text{5}}$ & 0 & 15$_{\pm \text{10}}$
& 20$_{\pm \text{15}}$ & 0 & 10$_{\pm \text{15}}$ & 0 & 5$_{\pm \text{5}}$\\

OpenVLA~\cite{Kim2024OpenVLAAO}
& 35$_{\pm \text{15}}$ & 30$_{\pm \text{10}}$ & 45$_{\pm \text{5}}$ & 0 & 45$_{\pm \text{10}}$
& 35$_{\pm \text{5}}$ & 0 & 30$_{\pm \text{10}}$ & 0 & 10$_{\pm \text{5}}$\\

$\pi_0$~\cite{Black20240AV}
& 50$_{\pm \text{5}}$ & 35$_{\pm \text{5}}$ & 55$_{\pm \text{15}}$ & 15$_{\pm \text{5}}$ & 75$_{\pm \text{5}}$
& 55$_{\pm \text{10}}$ & 25$_{\pm \text{5}}$ & 55$_{\pm \text{15}}$ & 15$_{\pm \text{5}}$ & 35$_{\pm \text{10}}$ \\

$\pi_{0.5}$~\cite{Intelligence202505AV}
& 65$_{\pm \text{5}}$ & 35$_{\pm \text{10}}$ & 60$_{\pm \text{10}}$ & 30$_{\pm \text{10}}$ & 75$_{\pm \text{5}}$ 
& 65$_{\pm \text{5}}$ & 25$_{\pm \text{10}}$ & 60$_{\pm \text{10}}$ & 20$_{\pm \text{5}}$ & 35$_{\pm \text{5}}$ \\

\midrule
\rowcolor{gray!15} 

$\pi_0$ + Registers 
& 75$_{\pm \text{10}}$ & 45$_{\pm \text{10}}$ & 75$_{\pm \text{5}}$ & 25$_{\pm \text{5}}$ & 75$_{\pm \text{10}}$
& 65$_{\pm \text{5}}$ & 30$_{\pm \text{15}}$ & 65$_{\pm \text{5}}$ & 25$_{\pm \text{5}}$ & 60$_{\pm \text{10}}$ \\

$\pi_0$ + Cropping
& 65$_{\pm \text{10}}$ & 40$_{\pm \text{15}}$ & 60$_{\pm \text{10}}$ & 25$_{\pm \text{10}}$ & 70$_{\pm \text{10}}$
& 70$_{\pm \text{5}}$ & 35$_{\pm \text{15}}$ & 65$_{\pm \text{10}}$ & 30$_{\pm \text{5}}$ & 50$_{\pm \text{15}}$ \\

\textbf{\textit{AtVLA} (Full)}
& \textbf{80}$_{\pm \text{10}}$ & \textbf{65}$_{\pm \text{5}}$ & \textbf{80}$_{\pm \text{5}}$ & \textbf{40}$_{\pm \text{5}}$ & \underline{75}$_{\pm \text{10}}$
& \textbf{80}$_{\pm \text{5}}$ & \textbf{50}$_{\pm \text{10}}$ & \textbf{70}$_{\pm \text{5}}$ & \textbf{35}$_{\pm \text{5}}$ & \textbf{70}$_{\pm \text{10}}$ \\

\bottomrule
\end{tabular}

\label{tab:realworld}

\end{table*}

\subsection{Attention-Guided Local Refinement}

When $U_t>\tau$, we derive a task-conditioned spatial map from attention matrices already produced during base action sampling. For layer $l$, we average the attention matrices over heads and a fixed subset of denoising steps, add the residual connection, and row-normalize:
\begin{equation}
    \widetilde{\mathbf{M}}^{\,l}
    =
    \operatorname{RowNorm}\left(
    \mathbf{I}+
    \frac{1}{|\mathcal{H}||\mathcal{Q}|}
    \sum_{m\in\mathcal{H}}\sum_{q\in\mathcal{Q}}
    \mathbf{M}^{\,l,m,q}\right)
\end{equation}
Attention rollout propagates token influence through the action expert:
\begin{equation}
    \mathbf{M}^{\mathrm{roll}}
    =
    \widetilde{\mathbf{M}}^{\,L}
    \widetilde{\mathbf{M}}^{\,L-1}\cdots
    \widetilde{\mathbf{M}}^{\,1}
\end{equation}
We average the entries from the first $h$ action-token rows to the base image-token columns and reshape them into a $16\times16$ saliency map $\mathbf{S}_t$. Register columns are excluded from spatial localization since registers provide global context, whereas only patch tokens retain a
fixed image-grid correspondence. The saliency map is bilinearly upsampled to the input resolution. We search a predefined set of square windows and select the region whose average internal attention is most distinctive from its surrounding ring:
\begin{equation}
\small
    b_t^{*} =
    \arg\max_{b\in\mathcal{B}}
    \left[
    \frac{1}{|b|}\sum_{\mathbf{u}\in b}S_t(\mathbf{u})
    -
    \frac{1}{|\rho(b)\setminus b|}
    \sum_{\mathbf{u}\in\rho(b)\setminus b}S_t(\mathbf{u})
    \right]
    \label{eq:crop-selection}
\end{equation}
where $\rho(b)$ is a context-expanded version of $b$. This contrastive criterion discourages both diffuse full-image crops and small windows centered on isolated attention peaks.

The selected region is cropped from $\mathbf{I}_t$, resized to
$224\times224$, and encoded by the same register-augmented vision encoder. The crop-side register outputs are discarded because the base registers already summarize global context; only the $N_c=256$ crop patch features $\mathbf{P}_t^{c}$ are retained. To preserve the crop's position in the original image, we compute
\begin{equation}
    \mathbf{e}_{b}
    =
    \operatorname{MLP}\left(
    \frac{x_1}{W},\frac{y_1}{H_I},
    \frac{x_2}{W},\frac{y_2}{H_I}\right)
\end{equation}
and add it to every projected crop token. The augmented prefix is
\begin{equation}
    \mathbf{C}_t^{+}
    =
    [\mathbf{C}_t^{b};
    \Pi(\mathbf{P}_t^{c})
    +\mathbf{1}_{N_c}\mathbf{e}_{b}^{\top}].
    \label{eq:aug-prefix}
\end{equation}
Only the newly appended crop segment is prefetched; the key--value cache of the base image, instruction, and state is reused. Action expert is then rerun under $\mathbf{C}_t^{+}$ to produce the refined action chunk.
Consequently, confident steps incur no crop encoding, while uncertain steps pay for one additional vision encoding and one augmented action generation.

\begin{figure*}
    \centering
    \includegraphics[width=1\linewidth]{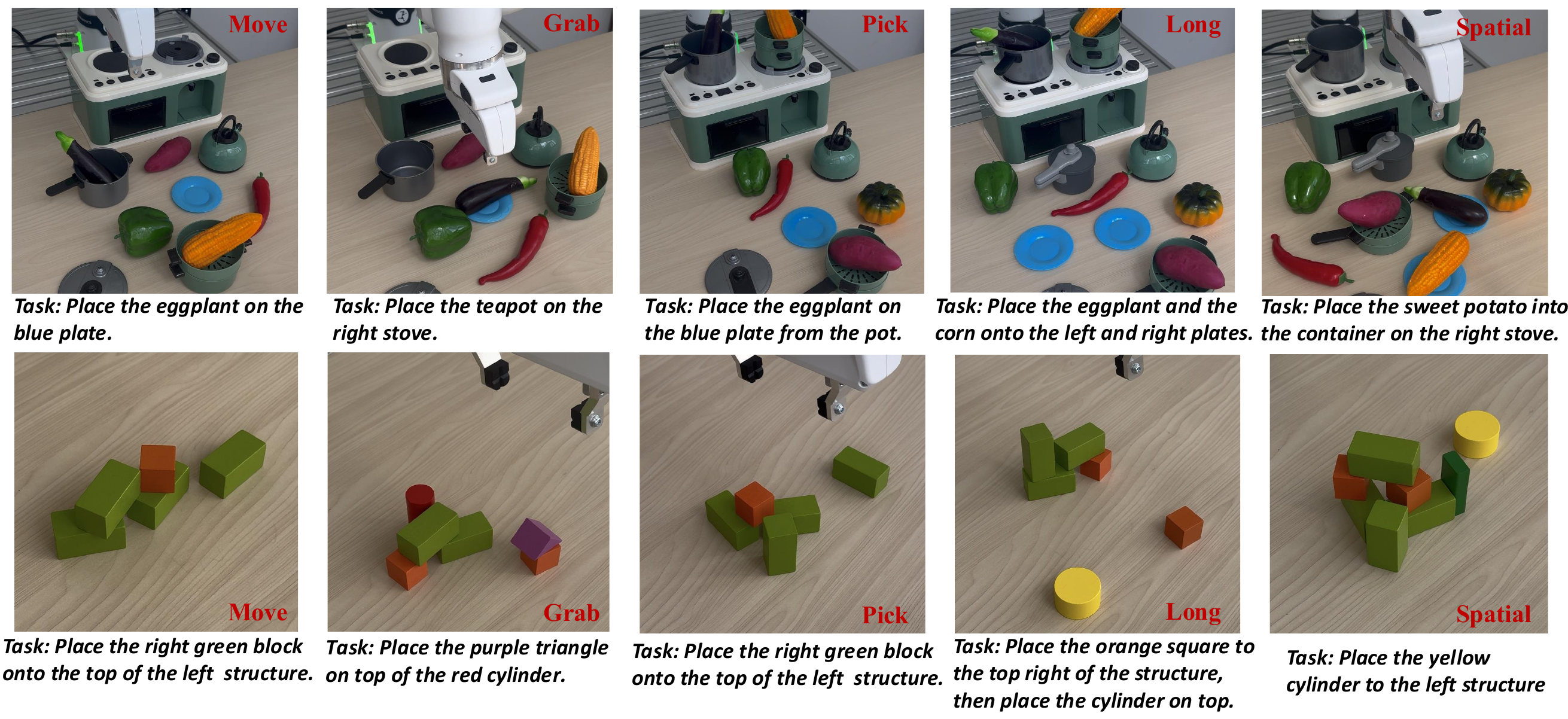}
    \vspace{-4mm}
    \caption{Real-world evaluation benchmark. The \textsc{Kitchen} suite (top) evaluates visual grounding and reasoning in cluttered household scenes, while \textsc{Building Blocks} (bottom) tests small-object stacking, spatial relations, and long-horizon construction.}
    \label{fig:realworld}
    \vspace{-2mm}
\end{figure*}

\subsection{Training Objective and Paradigm}
\label{sec:training}

Our model is initialized from a pretrained $\pi_0$ policy and retains its original vision--language backbone and action-generation formulation. We therefore use the standard $\pi_0$ action objective, denoted by $\mathcal{L}_{\pi_0}$, without modifying its action representation or generation process. Training focuses on integrating the newly introduced register tokens, crop-position encoding, and attention-guided localization while preserving the pretrained policy capabilities.

We adopt a three-stage post-training paradigm. First, we adapt the register tokens to embodied observations using the original action supervision, initially updating only the registers and subsequently unfreezing the last few layers of the visual encoder. Second, we introduce ground-truth crops and train the crop-position encoder and bounding-box predictor while keeping the pretrained visual representation fixed. Finally, we jointly optimize action prediction, crop localization, and action-conditioned attention grounding. During this stage, training progressively transitions from ground-truth crops to crops generated by model's own attention rollout, while crop dropout exposes the policy to both the base-only and crop-augmented inference paths. The final objective is
\begin{equation}
    \mathcal{L}
    =
    \mathcal{L}_{\pi_0}
    +
    \lambda_{\mathrm{cp}}\mathcal{L}_{\mathrm{cp}}
    +
    \lambda_{\mathrm{ag}}\mathcal{L}_{\mathrm{ag}},
    \label{eq:training-objective}
\end{equation}
where $\mathcal{L}_{\mathrm{cp}}$ supervises the predicted crop coordinates and $\mathcal{L}_{\mathrm{ag}}$ encourages action-to-image attention to concentrate inside the task-relevant object region. We use
$\lambda_{\mathrm{cp}}=0.1$ and $\lambda_{\mathrm{ag}}=1.0$. Detailed optimization schedules and parameter configurations are provided in the Appendix.

\section{Experiments}
\label{sec:experiments}

\subsection{Experimental Setup}
\label{sec:exp_setup}

\paragraph{Evaluation.}
We evaluate \textbf{\textit{AtVLA}} on two public manipulation benchmarks,
LIBERO~\cite{Liu2023LIBEROBK} and SimplerEnv~\cite{Li2024EvaluatingRR},
together with a real-world benchmark constructed on a Franka Research~3
robot. 


\textbf{\textit{LIBERO.}}
We evaluate on the four standard LIBERO suites:
\textsc{Spatial}, \textsc{Object}, \textsc{Goal}, and \textsc{Long-10},
which respectively assess spatial-relation generalization, object
generalization, instruction and goal understanding, and temporally
consistent execution in multi-stage tasks. We follow the standard
training and evaluation splits and report mean task success rates, with
standard deviations when available.

\textbf{\textit{SimplerEnv.}}
We evaluate the Google Robot suit of SimplerEnv under its official
protocol, covering \emph{Pick Coke Can}, \emph{Move Near}, and
\emph{Open/Close Drawer}. These tasks assess pose-robust grasping,
object-relative spatial reasoning, and articulated-object interaction.
We omit the WidowX suite since it requires BridgeData-specific
training and embodiment adaptation, which would confound visual
improvements with embodiment transfer; its tabletop settings are also
covered by our more challenging real-world benchmark.

\begin{figure*}
    \centering
    \includegraphics[width=1\linewidth]{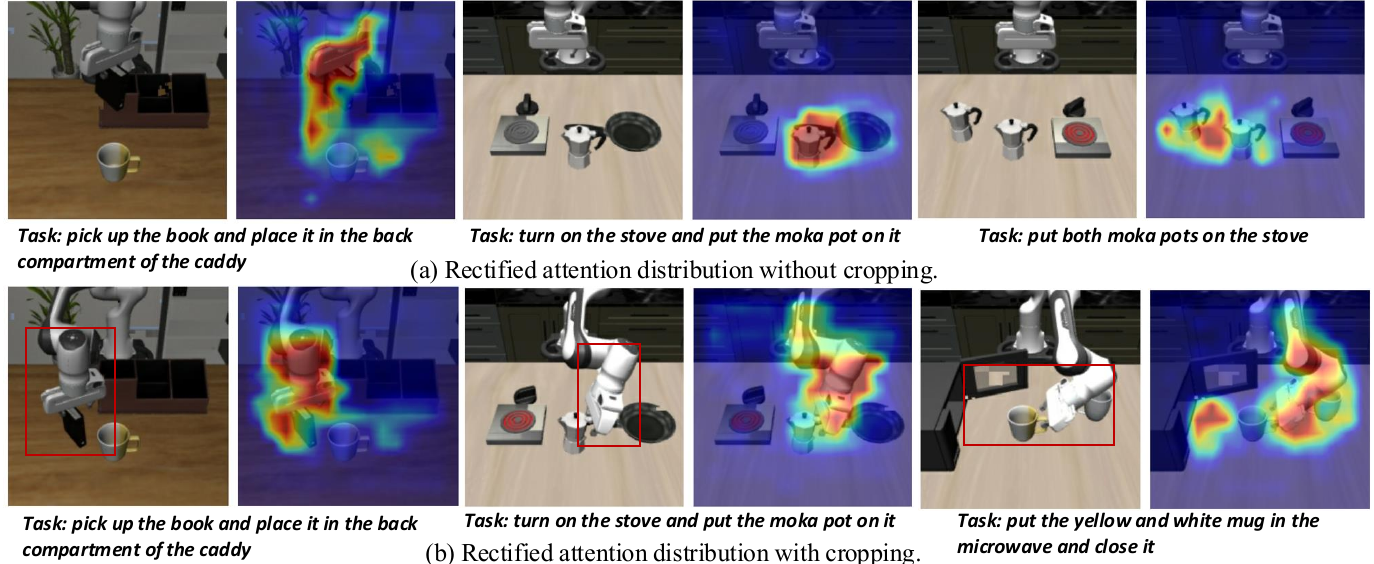}
    \caption{Qualitative effects of attention rectification and adaptive visual refinement. (a) Register-enhanced encoding produces clean attention aligned with task-relevant objects. (b) When uncertainty triggers refinement, attention rollout selects the interaction-critical region, whose high-resolution re-encoding sharpens perception around the target and contact area.}
    \label{fig:attention-cropping}
    \vspace{-2mm}
\end{figure*}

\noindent \textbf{\textit{Real-world evaluation.}}
Our real-world benchmark comprises two suites, \textsc{Kitchen} and
\textsc{Building Blocks}. \textsc{Kitchen} evaluates visual grounding,
contact-point localization, and spatial reasoning in cluttered household
scenes. \textsc{Building Blocks} emphasizes small-object grasping,
stacking, and structure construction, where similar appearances and
single-view depth ambiguity demand implicit 3D reasoning and precise
placement. Both suites include long-horizon tasks that test persistent
planning and behavioral consistency.

\paragraph{Model and implementation details.}
All \textbf{\textit{AtVLA}} variants are initialized from the same pretrained
$\pi_0$ checkpoint and retain its original action representation. The base policy contains a 3B PaliGemma model, comprising a SigLIP-So400m visual encoder and a Gemma-2B language backbone, together with a 300M-parameter action expert. Four 1152-dimensional register embeddings add only $4{,}608$ parameters, while the crop-position MLP and auxiliary bounding-box head are also lightweight relative to the backbone.

Both full images and cropped regions are resized to $224\times224$. We
use four register tokens and rank-$32$ LoRA adapters. Register adaptation
runs for 20K, followed by 10K steps of crop alignment and 40K
steps of joint refinement. Training is
distributed over eight NVIDIA RTX 6000 Ada GPUs.
All real-world policies are deployed on a NVIDIA RTX 4090 GPU with
24\,GB memory and Franka Research~3 robot with a single third-person Intel RealSense D435i camera.

\paragraph{Baselines and variants.}
We select representative generalist VLAs and
fine-tuning methods listed in Table.~\ref{tab:main_results}. For real-world evaluation, all locally evaluated methods use the same camera observation, task demonstrations, and are fine-tuned with comparable steps. We further compare three controlled $\pi_0$ variants: registers only, cropping only, and the complete model. These variants share the same initialization and training data, isolating the contribution of each component.

\begin{figure*}[htbp]
    \centering
    \includegraphics[width=1\linewidth]{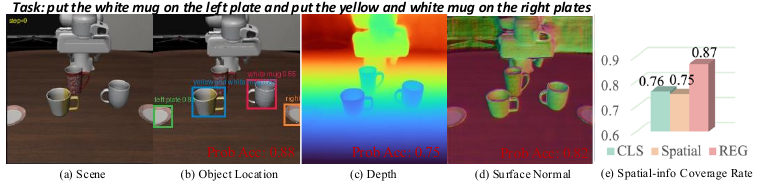}
    \caption{Probing embodied spatial information in visual tokens. Independent linear probes predict object location, depth, and surface normals from the CLS token, mean-pooled patch tokens, or register tokens; bars report normalized held-out scores. Register features consistently retain the richest task-relevant spatial information.}
    \label{fig:reg-info}
    \vspace{-2mm}
\end{figure*}

\subsection{Results and Analysis}

\paragraph{Benchmark evaluation.}
As shown in Table~\ref{tab:main_results}-\ref{tab:realworld}, \textbf{\textit{AtVLA}} achieves the best performance across all
four LIBERO suites, leads on two SimplerEnv tasks, and remains competitive on \emph{Open/Close Drawer}. It also obtains the
strongest performance across all real-world tasks. The gains are particularly evident on object-centric tasks, including LIBERO-\textsc{Object} and Kitchen \textsc{Grab}/\textsc{Pick}, as well as spatially demanding tasks LIBERO-\textsc{Spatial} and Building Blocks \textsc{Spatial}/\textsc{Grab}. Fig.~\ref{fig:attention-cropping} and the Appendix further show that \textbf{\textit{AtVLA}} consistently attends to task-relevant regions and invokes cropping near fine-grained interaction areas. These results confirm that rectified attention improves perception, while RoI refinement recovers local details required for precise manipulation.

Both controlled variants also improve over the original $\pi_0$. $\pi_0+\text{Registers}$ performs particularly well on spatial reasoning and object localization, consistent with its cleaner attention maps. $\pi_0+\text{Cropping}$ improves several fine-grained and long-horizon tasks, although its gains depend strongly on crop quality. We next analyze
the four questions introduced above.

\paragraph{Q1: Does attention rectification work in embodied policies, and can the repaired attention reliably guide cropping?}
Fig.~\ref{fig:artifacts} shows that adding register tokens largely removes the high-norm artifacts observed in Fig.~\ref{fig:artifacts} and redirects attention toward objects and regions relevant to action generation. The resulting maps also provide reliable localization signals for cropping. This demonstrates that attention artifacts can be corrected through embodied post-training, without additional vision-only pretraining or generic visual supervision. We also experimented with mixing ImageNet and CC3M datasets for separate visual adaptation; this introduced additional cost and noticeably disrupted capabilities acquired from robot demonstrations. We therefore train the registers entirely with embodied
data and action supervision.

\paragraph{Q2: How much embodied task information is captured by register tokens?}
We use linear probes to measure task-relevant spatial information contained in the CLS token, spatial patch features, and register tokens. As shown in Fig.~\ref{fig:reg-info}, registers consistently provide the strongest probe performance for object localization, depth, and surface-normal prediction, with all three scores exceeding $0.70$. This supports our hypothesis that embodied training introduces spatial and geometric information beyond the capacity of the original global token. Without dedicated storage slots, this information spills into patch tokens and produces attention artifacts; registers instead absorb and preserve it as structured context.

Removing the learned registers from $\pi_0+\text{Registers}$ causes a substantial degradation across all LIBERO suites, indicating that they do not merely suppress artifacts but retain task-relevant embodied knowledge.

\begin{table}[t]
\centering
\small
\caption{Ablation of the learned register tokens on LIBERO. Results are
success rates (\%); the final column reports the average task-level change.}
\label{tab:remove_registers}
\begin{tabular}{lccccc}
\toprule
Variant & Spatial & Object & Goal & Long-10 & Avg. $\Delta$ \\
\midrule
w/o REGs & 93.2 & 92.1 & 92.2 & 81.3 & $-4.45$ \\
\bottomrule
\end{tabular}
\vspace{-4mm}
\end{table}

\paragraph{Q3: Which tasks benefit from cropping?}
Beyond the expected gains in fine-grained perception, Tables~1--2 reveal several additional effects of cropping. First, the crops used by $\pi_0+\text{Cropping}$ are generated by an external VLM without embodied post-training. Qualitative inspection shows that these crops often fail to
isolate the interaction-critical region, which is consistent with their slightly lower performance than full \textbf{\textit{AtVLA}} and further highlights the importance of embodied supervision for learning decision-relevant visual grounding. Second, cropped observations improve LIBERO-\textsc{Long-10} and both real-world long-horizon categories. We attribute this gain to the additional visual context provided for subsequent decisions, suggesting that cropping can support not only local precision but also temporally extended planning. In contrast, the SimplerEnv \emph{Open/Close Drawer} scene is largely occupied by the cabinet, causing most crops to preserve nearly the entire original image rather than resolve the handle or contact point, and thus yielding suboptimal performance. Overall, additional
task-relevant visual context is beneficial, but the remaining results of $\pi_0+\text{Cropping}$ also show that inaccurate crops can distort the policy input, reduce decision accuracy, and increase action uncertainty when no clean attention signal is available to guide them.

\paragraph{Q4: What is the system cost of adaptive cropping?}
Let $C_{\mathrm{base}}$ denote one visual encoding and prefix prefill,
$C_{\mathrm{A}}$ one action-expert sampling pass, $C_{\mathrm{crop}}$ the
incremental crop encoding and prefill, $K$ the number of action samples,
$r$ the crop-trigger rate, and $N$ the number of replanning steps. The
approximate task-level computation is
\begin{align}
    C_{\pi_0}
    &=N(C_{\mathrm{base}}+C_{\mathrm{A}}),\\
    C_{\mathrm{AtVLA}}
    &=N\!\left(C_{\mathrm{base}}+K C_{\mathrm{A}}
      +r(C_{\mathrm{crop}}+C_{\mathrm{A}})\right).
\end{align}
The attention rollout itself adds little cost because it reuses attention computed during action sampling, while the base prefix KV cache is retained when crop tokens are appended. Under a representative setting of $K=4$ and $r\approx0.3$, and assuming that one expert pass and crop refinement
cost approximately $10$--$15\%$ and $20$--$25\%$ of a full $\pi_0$ replanning pass, respectively, AtVLA requires roughly $1.4$--$1.6\times$ the total computation of $\pi_0$ for the same number of steps. The exact latency and trigger-dependent cost are reported using measurements on the deployment RTX~4090.

\paragraph{Number of register tokens.}

\begin{figure}[htbp]
    \centering
    \includegraphics[width=1\linewidth]{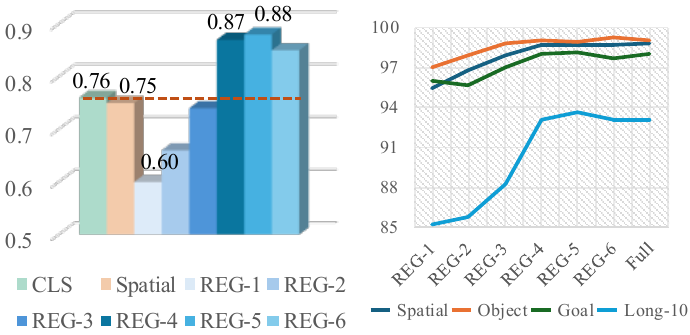}
    \caption{Effect of register-token capacity. \textbf{Left}: probe scores for the CLS token, patch tokens, and different numbers of registers. \textbf{Right}: success rates in LIBERO suites. }
    \label{fig:reg-num}
\end{figure}


As shown in Fig.~\ref{fig:reg-num}, using only a small number of register tokens can introduce additional interference rather than improve performance. We attribute this to incomplete absorption of the attention artifacts: the remaining corrupted patch features disrupt the coordination established between the vision encoder and the LLM backbone during embodied post-training, causing task performance to drop, in some cases even below the $\pi_0$ baseline. Performance saturates when the number of registers reaches four, while the amount of spatial information measured by linear probing also ceases to increase. We therefore adopt four register tokens as the smallest sufficient capacity.

\section{Conclusion}

We presented AtVLA, which enhances visual perception in VLA visual encoders by reorganizing embodied spatial information into register tokens and selectively restoring missing local detail through uncertainty-gated, attention-guided cropping. Across simulated and single-view real-world benchmarks, AtVLA substantially improves precise manipulation with modest additional computation.

\paragraph{}

\bibliography{aaai2027}

\begin{thebibliography}{31}
\providecommand{\natexlab}[1]{#1}

\bibitem[{Bendikas et~al.(2025)Bendikas, Dijkman, Peschl, Haresh, and Mazzaglia}]{Bendikas2025FocusingOW}
Bendikas, R.; Dijkman, D.; Peschl, M.; Haresh, S.; and Mazzaglia, P. 2025.
\newblock Focusing on What Matters: Object-Agent-centric Tokenization for Vision Language Action models.
\newblock \emph{ArXiv}, abs/2509.23655.

\bibitem[{Black et~al.(2024)Black, Brown, Driess, Esmail, Equi, Finn, Fusai, Groom, Hausman, Ichter, Jakubczak, Jones, Ke, Levine, Li-Bell, Mothukuri, Nair, Pertsch, Shi, Tanner, Vuong, Walling, Wang, and Zhilinsky}]{Black20240AV}
Black, K.; Brown, N.; Driess, D.; Esmail, A.; Equi, M.; Finn, C.; Fusai, N.; Groom, L.; Hausman, K.; Ichter, B.; Jakubczak, S.; Jones, T.; Ke, L.; Levine, S.; Li-Bell, A.; Mothukuri, M.; Nair, S.; Pertsch, K.; Shi, L.~X.; Tanner, J.; Vuong, Q.; Walling, A.; Wang, H.; and Zhilinsky, U. 2024.
\newblock $\pi$0: A Vision-Language-Action Flow Model for General Robot Control.
\newblock \emph{ArXiv}, abs/2410.24164.

\bibitem[{Brohan et~al.(2023)Brohan, Brown, Carbajal, Chebotar, Chen, Choromanski, Ding, Driess, Dubey, Finn, Florence, Fu, Arenas, Gopalakrishnan, Han, Hausman, Herzog, Hsu, Ichter, Irpan, Joshi, Julian, Kalashnikov, Kuang, Leal, Lee, Lee, Levine, Lu, Michalewski, Mordatch, Pertsch, Rao, Reymann, Ryoo, Salazar, Sanketi, Sermanet, Singh, Singh, Soricut, Tran, Vanhoucke, Vuong, Wahid, Welker, Wohlhart, Wu, Xia, Xiao, Xu, Xu, Yu, and Zitkovich}]{brohan2023rt2visionlanguageactionmodelstransfer}
Brohan, A.; Brown, N.; Carbajal, J.; Chebotar, Y.; Chen, X.; Choromanski, K.; Ding, T.; Driess, D.; Dubey, A.; Finn, C.; Florence, P.; Fu, C.; Arenas, M.~G.; Gopalakrishnan, K.; Han, K.; Hausman, K.; Herzog, A.; Hsu, J.; Ichter, B.; Irpan, A.; Joshi, N.; Julian, R.; Kalashnikov, D.; Kuang, Y.; Leal, I.; Lee, L.; Lee, T.-W.~E.; Levine, S.; Lu, Y.; Michalewski, H.; Mordatch, I.; Pertsch, K.; Rao, K.; Reymann, K.; Ryoo, M.; Salazar, G.; Sanketi, P.; Sermanet, P.; Singh, J.; Singh, A.; Soricut, R.; Tran, H.; Vanhoucke, V.; Vuong, Q.; Wahid, A.; Welker, S.; Wohlhart, P.; Wu, J.; Xia, F.; Xiao, T.; Xu, P.; Xu, S.; Yu, T.; and Zitkovich, B. 2023.
\newblock RT-2: Vision-Language-Action Models Transfer Web Knowledge to Robotic Control.
\newblock arXiv:2307.15818.

\bibitem[{Carvalho, Dias, and Martins(2026)}]{carvalho2026cropvlmlearningzoomfinegrained}
Carvalho, M.; Dias, H.; and Martins, B. 2026.
\newblock CropVLM: Learning to Zoom for Fine-Grained Vision-Language Perception.
\newblock arXiv:2511.19820.

\bibitem[{Darcet et~al.(2023)Darcet, Oquab, Mairal, and Bojanowski}]{Darcet2023VisionTN}
Darcet, T.; Oquab, M.; Mairal, J.; and Bojanowski, P. 2023.
\newblock Vision Transformers Need Registers.
\newblock \emph{ArXiv}, abs/2309.16588.

\bibitem[{Deng et~al.(2025)Deng, Yan, Wei, Ma, Yang, Chen, Zhang, Yang, Zhang, Cui, Zhang, and Wang}]{Deng2025GraspVLAAG}
Deng, S.; Yan, M.; Wei, S.; Ma, H.-L.; Yang, Y.; Chen, J.; Zhang, Z.; Yang, T.; Zhang, X.; Cui, H.; Zhang, Z.; and Wang, H. 2025.
\newblock GraspVLA: a Grasping Foundation Model Pre-trained on Billion-scale Synthetic Action Data.
\newblock \emph{ArXiv}, abs/2505.03233.

\bibitem[{Hanyu et~al.(2025)Hanyu, Chung, Le, Nguyen, Ikebe, Gunderman, Minh, Vo, Kieu, Yamazaki, Rainwater, Nguyen, and Le}]{Hanyu2025SlotVLATM}
Hanyu, T.; Chung, N.; Le, H.; Nguyen, T.; Ikebe, Y.; Gunderman, A.; Minh, D. N.~H.; Vo, K.~T.; Kieu, T.; Yamazaki, K.; Rainwater, C.; Nguyen, A.; and Le, N. 2025.
\newblock SlotVLA: Towards Modeling of Object-Relation Representations in Robotic Manipulation.
\newblock \emph{ArXiv}, abs/2511.06754.

\bibitem[{Intelligence et~al.(2025)Intelligence, Black, Brown, Darpinian, Dhabalia, Driess, Esmail, Equi, Finn, Fusai, Galliker, Ghosh, Groom, Hausman, Ichter, Jakubczak, Jones, Ke, LeBlanc, Levine, Li-Bell, Mothukuri, Nair, Pertsch, Ren, Shi, Smith, Springenberg, Stachowicz, Tanner, Vuong, Walke, Walling, Wang, Yu, and Zhilinsky}]{Intelligence202505AV}
Intelligence, P.; Black, K.; Brown, N.; Darpinian, J.; Dhabalia, K.; Driess, D.; Esmail, A.; Equi, M.; Finn, C.; Fusai, N.; Galliker, M.~Y.; Ghosh, D.; Groom, L.; Hausman, K.; Ichter, B.; Jakubczak, S.; Jones, T.; Ke, L.; LeBlanc, D.; Levine, S.; Li-Bell, A.; Mothukuri, M.; Nair, S.; Pertsch, K.; Ren, A.~Z.; Shi, L.~X.; Smith, L.; Springenberg, J.~T.; Stachowicz, K.; Tanner, J.; Vuong, Q.; Walke, H.~R.; Walling, A.; Wang, H.; Yu, L.; and Zhilinsky, U. 2025.
\newblock $\pi$0.5: a Vision-Language-Action Model with Open-World Generalization.
\newblock \emph{ArXiv}, abs/2504.16054.

\bibitem[{Jiang et~al.(2024)Jiang, Chen, Zhu, Luo, Shen, and Yang}]{Jiang2024DevilsIM}
Jiang, Z.; Chen, J.; Zhu, B.; Luo, T.; Shen, Y.; and Yang, X. 2024.
\newblock Devils in Middle Layers of Large Vision-Language Models: Interpreting, Detecting and Mitigating Object Hallucinations via Attention Lens.
\newblock \emph{2025 IEEE/CVF Conference on Computer Vision and Pattern Recognition (CVPR)}, 25004--25014.

\bibitem[{Kachaev et~al.(2025)Kachaev, Kolosov, Zelezetsky, Kovalev, and Panov}]{Kachaev2025DontBY}
Kachaev, N.; Kolosov, M.; Zelezetsky, D.; Kovalev, A.~K.; and Panov, A.~I. 2025.
\newblock Don't Blind Your VLA: Aligning Visual Representations for OOD Generalization.
\newblock \emph{ArXiv}, abs/2510.25616.

\bibitem[{Kaduri, Bagon, and Dekel(2024)}]{Kaduri2024WhatsIT}
Kaduri, O.; Bagon, S.; and Dekel, T. 2024.
\newblock What’s in the Imageƒ A Deep-Dive into the Vision of Vision Language Models.
\newblock \emph{2025 IEEE/CVF Conference on Computer Vision and Pattern Recognition (CVPR)}, 14549--14558.

\bibitem[{Kim, Finn, and Liang(2025)}]{Kim2025FineTuningVM}
Kim, M.~J.; Finn, C.; and Liang, P. 2025.
\newblock Fine-Tuning Vision-Language-Action Models: Optimizing Speed and Success.
\newblock \emph{ArXiv}, abs/2502.19645.

\bibitem[{Kim et~al.(2024)Kim, Pertsch, Karamcheti, Xiao, Balakrishna, Nair, Rafailov, Foster, Lam, Sanketi, Vuong, Kollar, Burchfiel, Tedrake, Sadigh, Levine, Liang, and Finn}]{Kim2024OpenVLAAO}
Kim, M.~J.; Pertsch, K.; Karamcheti, S.; Xiao, T.; Balakrishna, A.; Nair, S.; Rafailov, R.; Foster, E.~P.; Lam, G.; Sanketi, P.~R.; Vuong, Q.; Kollar, T.; Burchfiel, B.; Tedrake, R.; Sadigh, D.; Levine, S.; Liang, P.; and Finn, C. 2024.
\newblock OpenVLA: An Open-Source Vision-Language-Action Model.
\newblock \emph{ArXiv}, abs/2406.09246.

\bibitem[{Li et~al.(2025{\natexlab{a}})Li, Wen, Peng, Peng, Feng, and Zhu}]{Li2025PointVLAIT}
Li, C.; Wen, J.; Peng, Y.; Peng, Y.; Feng, F.; and Zhu, Y. 2025{\natexlab{a}}.
\newblock PointVLA: Injecting the 3D World into Vision-Language-Action Models.
\newblock \emph{ArXiv}, abs/2503.07511.

\bibitem[{Li et~al.(2025{\natexlab{b}})Li, Heng, Liu, Shen, Gu, Liu, Chen, Han, Zhang, Tang, Zhang, and Dong}]{pmlr-v305-li25g}
Li, X.; Heng, L.; Liu, J.; Shen, Y.; Gu, C.; Liu, Z.; Chen, H.; Han, N.; Zhang, R.; Tang, H.; Zhang, S.; and Dong, H. 2025{\natexlab{b}}.
\newblock 3DS-VLA: A 3D Spatial-Aware Vision Language Action Model for Robust Multi-Task Manipulation.
\newblock In Lim, J.; Song, S.; and Park, H.-W., eds., \emph{Proceedings of The 9th Conference on Robot Learning}, volume 305 of \emph{Proceedings of Machine Learning Research}, 2344--2359. PMLR.

\bibitem[{Li et~al.(2024{\natexlab{a}})Li, Hsu, Gu, Pertsch, Mees, Walke, Fu, Lunawat, Sieh, Kirmani, Levine, Wu, Finn, Su, Vuong, and Xiao}]{Li2024EvaluatingRR}
Li, X.; Hsu, K.; Gu, J.; Pertsch, K.; Mees, O.; Walke, H.~R.; Fu, C.; Lunawat, I.; Sieh, I.; Kirmani, S.; Levine, S.; Wu, J.; Finn, C.; Su, H.; Vuong, Q.~H.; and Xiao, T. 2024{\natexlab{a}}.
\newblock Evaluating Real-World Robot Manipulation Policies in Simulation.
\newblock In \emph{Conference on Robot Learning}.

\bibitem[{Li et~al.(2024{\natexlab{b}})Li, Li, Qian, Liu, Wang, Liu, Kang, Ma, Wang, Guo, Kong, Zhang, and Liu}]{Li2024WhatMI}
Li, X.; Li, P.; Qian, L.; Liu, M.; Wang, D.; Liu, J.; Kang, B.; Ma, X.; Wang, X.; Guo, D.; Kong, T.; Zhang, H.; and Liu, H. 2024{\natexlab{b}}.
\newblock What matters in building vision–language–action models for generalist robots.
\newblock \emph{Nature Machine Intelligence}, 8: 158 -- 172.

\bibitem[{Li et~al.(2024{\natexlab{c}})Li, Ren, Yang, Zhao, Wu, Xu, Bai, and Zhao}]{Li2024VIPVI}
Li, Z.; Ren, L.; Yang, J.; Zhao, Y.; Wu, X.; Xu, Z.; Bai, X.; and Zhao, H. 2024{\natexlab{c}}.
\newblock VIP: Vision Instructed Pre-training for Robotic Manipulation.

\bibitem[{Liu et~al.(2023)Liu, Zhu, Gao, Feng, Liu, Zhu, and Stone}]{Liu2023LIBEROBK}
Liu, B.; Zhu, Y.; Gao, C.; Feng, Y.; Liu, Q.; Zhu, Y.; and Stone, P. 2023.
\newblock LIBERO: Benchmarking Knowledge Transfer for Lifelong Robot Learning.
\newblock \emph{ArXiv}, abs/2306.03310.

\bibitem[{Liu et~al.(2024)Liu, Dong, Rao, Zhou, and Lu}]{Liu2024ChainofSpotIR}
Liu, Z.; Dong, Y.; Rao, Y.; Zhou, J.; and Lu, J. 2024.
\newblock Chain-of-Spot: Interactive Reasoning Improves Large Vision-Language Models.
\newblock \emph{ArXiv}, abs/2403.12966.

\bibitem[{Qu et~al.(2025)Qu, Song, Chen, Yao, Ye, Ding, Wang, Gu, Zhao, Wang, and Li}]{Qu2025SpatialVLAES}
Qu, D.; Song, H.; Chen, Q.; Yao, Y.; Ye, X.; Ding, Y.; Wang, Z.; Gu, J.; Zhao, B.; Wang, D.; and Li, X. 2025.
\newblock SpatialVLA: Exploring Spatial Representations for Visual-Language-Action Model.
\newblock \emph{ArXiv}, abs/2501.15830.

\bibitem[{Shen et~al.(2024)Shen, Zhao, Zhao, Xu, Zhang, Zhu, and Yin}]{Shen2024ZoomEyeEM}
Shen, H.; Zhao, K.; Zhao, T.; Xu, R.; Zhang, Z.; Zhu, M.; and Yin, J. 2024.
\newblock ZoomEye: Enhancing Multimodal LLMs with Human-Like Zooming Capabilities through Tree-Based Image Exploration.
\newblock In \emph{Conference on Empirical Methods in Natural Language Processing}.

\bibitem[{Song et~al.(2025)Song, Zhou, Zhao, Chen, Ding, Yan, Huang, Tang, Wang, and Li}]{Song2025ReconVLARV}
Song, W.; Zhou, Z.; Zhao, H.; Chen, J.; Ding, P.; Yan, H.; Huang, Y.; Tang, F.; Wang, D.; and Li, H. 2025.
\newblock ReconVLA: Reconstructive Vision-Language-Action Model as Effective Robot Perceiver.
\newblock In \emph{AAAI Conference on Artificial Intelligence}.

\bibitem[{Team et~al.(2024)Team, Ghosh, Walke, Pertsch, Black, Mees, Dasari, Hejna, Kreiman, Xu, Luo, Tan, Sanketi, Vuong, Xiao, Sadigh, Finn, and Levine}]{Team2024OctoAO}
Team, O.~M.; Ghosh, D.; Walke, H.~R.; Pertsch, K.; Black, K.; Mees, O.; Dasari, S.; Hejna, J.; Kreiman, T.; Xu, C.; Luo, J.; Tan, Y.~L.; Sanketi, P.~R.; Vuong, Q.; Xiao, T.; Sadigh, D.; Finn, C.; and Levine, S. 2024.
\newblock Octo: An Open-Source Generalist Robot Policy.
\newblock \emph{ArXiv}, abs/2405.12213.

\bibitem[{Wu et~al.(2026)Wu, Fan, Liao, Jiang, Yang, Luo, Wu, Zheng, and Loy}]{Wu2026VLANeXtRF}
Wu, X.-M.; Fan, B.; Liao, K.; Jiang, J.-J.; Yang, R.; Luo, Y.; Wu, Z.; Zheng, W.; and Loy, C.~C. 2026.
\newblock VLANeXt: Recipes for Building Strong VLA Models.
\newblock \emph{ArXiv}, abs/2602.18532.

\bibitem[{Zawalski et~al.(2024)Zawalski, Chen, Pertsch, Mees, Finn, and Levine}]{Zawalski2024RoboticCV}
Zawalski, M.; Chen, W.; Pertsch, K.; Mees, O.; Finn, C.; and Levine, S. 2024.
\newblock Robotic Control via Embodied Chain-of-Thought Reasoning.
\newblock In \emph{Conference on Robot Learning}.

\bibitem[{Zhang et~al.(2024)Zhang, Khayatkhoei, Chhikara, and Ilievski}]{zhang2024perceivingsmallvisualdetails}
Zhang, J.; Khayatkhoei, M.; Chhikara, P.; and Ilievski, F. 2024.
\newblock Towards Perceiving Small Visual Details in Zero-shot Visual Question Answering with Multimodal LLMs.
\newblock arXiv:2310.16033.

\bibitem[{Zhang et~al.(2025)Zhang, Khayatkhoei, Chhikara, and Ilievski}]{Zhang2025MLLMsKW}
Zhang, J.; Khayatkhoei, M.; Chhikara, P.; and Ilievski, F. 2025.
\newblock MLLMs Know Where to Look: Training-free Perception of Small Visual Details with Multimodal LLMs.
\newblock \emph{ArXiv}, abs/2502.17422.

\bibitem[{Zhao et~al.(2025)Zhao, Lu, Kim, Fu, Zhang, Wu, Li, Ma, Han, Finn, Handa, Liu, Xiang, Wetzstein, and Lin}]{Zhao2025CoTVLAVC}
Zhao, Q.; Lu, Y.; Kim, M.~J.; Fu, Z.; Zhang, Z.; Wu, Y.; Li, Z.; Ma, Q.; Han, S.; Finn, C.; Handa, A.; Liu, M.-Y.; Xiang, D.; Wetzstein, G.; and Lin, T.-Y. 2025.
\newblock CoT-VLA: Visual Chain-of-Thought Reasoning for Vision-Language-Action Models.
\newblock \emph{2025 IEEE/CVF Conference on Computer Vision and Pattern Recognition (CVPR)}, 1702--1713.

\bibitem[{Zheng et~al.(2024)Zheng, Liang, Huang, Gao, Daum'e, Kolobov, Huang, and Yang}]{Zheng2024TraceVLAVT}
Zheng, R.; Liang, Y.; Huang, S.; Gao, J.; Daum'e, H.; Kolobov, A.; Huang, F.; and Yang, J. 2024.
\newblock TraceVLA: Visual Trace Prompting Enhances Spatial-Temporal Awareness for Generalist Robotic Policies.
\newblock \emph{ArXiv}, abs/2412.10345.

\bibitem[{Zhong et~al.(2025)Zhong, Rosenthal, Sicking, Hüger, Bagdonat, Gottschalk, and Schwinn}]{zhong2025focusinternalmllmrepresentations}
Zhong, L.; Rosenthal, F.; Sicking, J.; Hüger, F.; Bagdonat, T.; Gottschalk, H.; and Schwinn, L. 2025.
\newblock FOCUS: Internal MLLM Representations for Efficient Fine-Grained Visual Question Answering.
\newblock arXiv:2506.21710.

\end{thebibliography}

\appendix

\section{Training Data}
\label{app:training-data}

All reported models are post-trained exclusively on robot demonstrations. For LIBERO, we train a single unified policy on the mixture of the four
standard suites: \textsc{Spatial}, \textsc{Object}, \textsc{Goal}, and
\textsc{Long-10}. Each suite contains ten tasks with 50 expert
demonstrations per task, resulting in 40 tasks and 2,000 trajectories in
total. 

Each demonstration contains RGB observations, a language instruction,
proprioceptive states, and continuous action sequences. For crop-related
training, each observation is additionally associated with an annotated
task-relevant region used to construct ground-truth crops and supervise
action-conditioned attention. An episode-disjoint subset
of the training demonstrations is held out for uncertainty-threshold
calibration.

\section{Training Paradigm}
\label{app:training}

Rather than training the complete policy from scratch, we adopt a staged
post-training procedure on top of pretrained $\pi_0$. The design follows
two principles. First, newly introduced visual components are aligned
before being jointly optimized with the policy, preventing unstable
feature shifts from disrupting the pretrained vision--language
representation. Second, the crop-input distribution is progressively
aligned with inference, where crop regions are generated from the model's
own action-conditioned attention rather than from annotations.

Let $\mathcal{L}_{\pi_0}$ denote the original action-training objective of
$\pi_0$. We retain this objective throughout all stages. During joint
refinement, we additionally supervise the action-to-image attention:
\begin{equation}
    \mathcal{L}_{\mathrm{ag}}
    =
    -\log
    \frac{
        \sum_{p\in\Omega(\mathbf{b}^{\mathrm{gt}})}
        A_p+\epsilon
    }{
        \sum_{p}A_p+\epsilon
    },
    \label{eq:ag-loss}
\end{equation}
where $A_p$ denotes the aggregated attention assigned by the near-term
action tokens to base-image patch $p$, and
$\Omega(\mathbf{b}^{\mathrm{gt}})$ is the set of patch centers inside the
annotated task-relevant region. This objective encourages the attention
used for crop extraction to concentrate on regions relevant to the
current manipulation decision.

\subsection{Stage 0: Register Adaptation}

The first stage adapts the randomly initialized register tokens to
embodied visual observations. All training samples use the base view only,
and the crop branch is disabled. The pretrained PaliGemma backbone, visual
projector, and action expert remain frozen. The standard
$\mathcal{L}_{\pi_0}$ objective propagates task-level supervision through
the frozen policy to the trainable visual parameters.

Stage 0 spans 20K optimization steps and is divided into two consecutive
phases. During the first 4K steps, only the four register embeddings are
updated:
\begin{equation}
    \min_{\theta_{\mathrm{reg}}}
    \mathbb{E}_{(\mathbf{o},\ell,\mathbf{s},\mathbf{a})
    \sim\mathcal{D}}
    \left[
        \mathcal{L}_{\pi_0}
    \right].
\end{equation}
This warm-up allows the registers to acquire useful global-information
storage behavior without immediately perturbing the pretrained image
features.

For the remaining 16K steps, the last four SigLIP transformer blocks are
additionally unfrozen:
\begin{equation}
    \min_{\theta_{\mathrm{reg}},
         \theta_{\mathrm{vis}}^{\mathrm{last}}}
    \mathbb{E}_{\mathcal{D}}
    \left[
        \mathcal{L}_{\pi_0}
    \right].
\end{equation}
The earlier visual blocks and all downstream pretrained components remain
fixed. This gradual unfreezing enables the upper visual layers to
reorganize embodied global information into the register slots while
limiting drift in low-level visual representations. We use learning rates
of $1\times10^{-4}$ for the register-only phase and
$2\times10^{-5}$ after unfreezing the final visual blocks.

\subsection{Stage 1: Ground-Truth Crop Alignment}

After register adaptation, we introduce the crop-augmented input format.
Each training sample contains the base observation together with a
high-resolution crop extracted from the annotated task-relevant region.
Ground-truth crops are used exclusively in this stage, and crop dropout is
disabled.

The complete SigLIP encoder, register embeddings, visual projector, and
pretrained backbone weights remain frozen. We optimize the crop-position
encoder and rank-$32$ LoRA adapters inserted into the language and action
components using the original action objective:
\begin{equation}
    \mathcal{L}_{\mathrm{align}}
    =
    \mathcal{L}_{\pi_0}.
    \label{eq:stage1-loss}
\end{equation}

This stage teaches the policy to interpret the additional crop tokens and
their position in the original image without altering the
register-specialized visual encoder. The crop-position encoder maps the
normalized crop coordinates
$(x_1/W,y_1/H,x_2/W,y_2/H)$ to the multimodal hidden dimension and adds the
resulting embedding to every crop-view token. We train this stage for 10K
steps with a learning rate of $1\times10^{-4}$.

\subsection{Action-Conditioned Attention Grounding}

The crop used at inference is generated from action-to-image attention.
Optimizing only the action objective does not guarantee that this
attention identifies the task-relevant interaction region; it may instead
concentrate on visually salient backgrounds or persistent positional
sinks. We therefore activate $\mathcal{L}_{\mathrm{ag}}$ from the first
step of the final joint-training stage.

Attention grounding is not trained as an isolated continuation stage.
Instead, action prediction, visual adaptation, and attention calibration
co-adapt throughout the same optimization trajectory, avoiding abrupt
changes in the objective or learning-rate schedule.

\subsection{Stage 2: Joint Refinement and Inference Alignment}

The final stage jointly adapts the complete AtVLA policy. The base weights
of SigLIP, Gemma, and the action expert remain frozen, while rank-$32$ LoRA
adapters in all three components are trainable. The register embeddings,
visual projector, and crop-position encoder are optimized directly.

The joint objective is
\begin{equation}
    \mathcal{L}_{\mathrm{joint}}
    =
    \mathcal{L}_{\pi_0}
    +
    \lambda_{\mathrm{ag}}
    \mathcal{L}_{\mathrm{ag}},
    \qquad
    \lambda_{\mathrm{ag}}=1.0.
    \label{eq:joint-loss}
\end{equation}
The original action objective preserves control quality, while
$\mathcal{L}_{\mathrm{ag}}$ calibrates the internal attention used for
local refinement.

To reduce the mismatch between training with annotated crops and
inference with attention-derived crops, we employ a three-phase crop
curriculum over the 40K-step joint-training schedule. Let $\alpha$ denote
the probability of using a rollout-derived crop instead of a
ground-truth crop.

\paragraph{Stable alignment phase.}
During the first 40\% of joint training, all crops are extracted from
ground-truth regions:
\begin{equation}
    \alpha=0,
    \qquad
    p_{\mathrm{drop}}=0.
\end{equation}
This phase stabilizes multimodal fusion and prevents inaccurate early
attention maps from corrupting crop inputs.

\paragraph{Mixed transition phase.}
From 40\% to 90\% of joint training, $\alpha$ is linearly increased from
$0$ to $1$. For each non-dropped sample, the crop source is selected as
\begin{equation}
    \mathbf{b}^{\mathrm{crop}}
    =
    \begin{cases}
        \mathbf{b}^{\mathrm{roll}},
        & u < \alpha
          \ \text{and}\
          \operatorname{Valid}
          (\mathbf{b}^{\mathrm{roll}}),\\
        \mathbf{b}^{\mathrm{gt}},
        & u \geq \alpha,\\
        \varnothing,
        & u < \alpha
          \ \text{and}\
          \neg\operatorname{Valid}
          (\mathbf{b}^{\mathrm{roll}}),
    \end{cases}
    \qquad
    u\sim\mathcal{U}(0,1).
    \label{eq:crop-curriculum}
\end{equation}
Here, $\varnothing$ denotes the base-only input. Thus, an invalid rollout
does not fall back to a ground-truth crop; it skips crop refinement,
matching the failure-handling behavior used at deployment.

Over the same interval, the crop-dropout probability is linearly increased
from $0$ to $0.3$. Crop dropout removes the complete crop-view segment,
forcing the policy to retain strong performance under the base-only input
used by the confident inference path.

\paragraph{Inference-alignment phase.}
During the final 10\% of training, all valid non-dropped crops are obtained
from attention rollout:
\begin{equation}
    \alpha=1,
    \qquad
    p_{\mathrm{drop}}=0.3.
\end{equation}
If rollout localization is invalid, the crop segment is omitted and the
sample follows the base-only path. The policy is therefore optimized under
the same crop-generation and failure-handling mechanisms used at
deployment.

\subsection{Optimization Details}

Stage 2 is trained for 40K steps with AdamW, a batch size of 4, and a
constant learning rate of $1\times10^{-4}$. Together with the 20K-step
register adaptation and 10K-step crop alignment, the complete post-training
procedure contains 70K gradient updates.

Both base images and crop regions are resized to $224\times224$. Training
is distributed over eight NVIDIA RTX 6000 Ada GPUs. The same pretrained
$\pi_0$ action representation, action normalization, prediction horizon,
and flow-matching objective are retained across all stages. The procedure
therefore introduces visual specialization and adaptive re-examination
without changing the underlying action space or deployment interface of
the base policy.

\section{Uncertainty-Gated Refinement}
\label{app:uncertainty}

At each action-chunk replanning step, the base visual and language prefix
is computed once and stored in a reusable key--value cache. Conditioned on
this shared prefix, the action expert generates $K=4$ action chunks from
independent Gaussian initializations. We use the same ten-step Euler solver
and noise schedule for all samples.

Let $\mathbf{A}^{(k)}_{1:H}$ denote the $k$-th sampled action chunk, where
$H$ is the native action horizon of the underlying $\pi_0$ checkpoint.
Uncertainty is computed over the first $h$ actions, where $h$ equals the
number of low-level commands executed before the next policy query. Thus,
the uncertainty estimate always covers exactly the near-term portion of
the chunk that influences closed-loop execution.

Using the translational action dimensions
$\mathcal{D}_{\mathrm{tr}}=\{\Delta x,\Delta y,\Delta z\}$, we compute
\begin{equation}
    U_t
    =
    \frac{1}{h|\mathcal{D}_{\mathrm{tr}}|}
    \sum_{j=1}^{h}
    \sum_{d\in\mathcal{D}_{\mathrm{tr}}}
    \sqrt{
        \frac{1}{K-1}
        \sum_{k=1}^{K}
        \left(
            A^{(k)}_{j,d}
            -
            \overline{A}_{j,d}
        \right)^2
    },
    \label{eq:uncertainty-app}
\end{equation}
where
\begin{equation}
    \overline{A}_{j,d}
    =
    \frac{1}{K}
    \sum_{k=1}^{K}
    A^{(k)}_{j,d}.
\end{equation}

When $U_t\leq\tau$, the mean sampled action chunk is executed directly.
When $U_t>\tau$, attention-guided crop refinement is attempted. The
refined path performs one additional crop encoding and one additional
action generation rather than repeating all $K$ samples.

\subsection{Threshold Calibration}

The uncertainty scale depends on the embodiment-specific action
normalization and control interface. We therefore calibrate one threshold
per embodiment rather than using a single threshold across robots. In
particular, the same threshold is shared across all four LIBERO suites,
while separate thresholds are used for the Google Robot and Franka
embodiments.

For each sample in the episode-disjoint calibration set, we evaluate both
the base and crop-augmented paths and measure the reduction in near-term
translational action error:
\begin{equation}
    \Delta e_t
    =
    e_t^{\mathrm{base}}
    -
    e_t^{\mathrm{crop}}.
\end{equation}
We select the threshold that maximizes the expected benefit of refinement
subject to an approximate 30\% trigger-rate budget:
\begin{equation}
    \tau^{*}
    =
    \arg\max_{\tau}
    \;
    \mathbb{E}
    \left[
        \Delta e_t
        \mathbb{1}(U_t>\tau)
    \right],
    \qquad
    \Pr(U_t>\tau)\leq0.3.
    \label{eq:threshold-selection}
\end{equation}
In practice, the 70th percentile of the validation uncertainty
distribution provides the initial threshold, which is then adjusted using
the measured crop benefit.

\section{Attention Rollout and Crop Extraction}
\label{app:crop-details}

\subsection{Rollout Aggregation}

We collect action-to-prefix attention from all 18 joint-attention layers
of the action expert. Attention matrices are averaged over all heads and
over Euler denoising steps
\begin{equation}
    \mathcal{Q}=\{3,6,9\}.
\end{equation}
These steps provide complementary early, intermediate, and late
action-conditioned signals without storing attention from the complete
denoising trajectory.

For layer $l$, we compute
\begin{equation}
    \widetilde{\mathbf{M}}_l
    =
    \operatorname{RowNorm}
    \left(
        \mathbf{I}
        +
        \frac{1}{|\mathcal{H}||\mathcal{Q}|}
        \sum_{m\in\mathcal{H}}
        \sum_{q\in\mathcal{Q}}
        \mathbf{M}_{l,m,q}
    \right),
    \label{eq:rollout-layer-app}
\end{equation}
where $\mathcal{H}$ denotes the set of attention heads. Attention is then
propagated across layers:
\begin{equation}
    \mathbf{M}_{\mathrm{roll}}
    =
    \widetilde{\mathbf{M}}_L
    \widetilde{\mathbf{M}}_{L-1}
    \cdots
    \widetilde{\mathbf{M}}_1.
    \label{eq:rollout-app}
\end{equation}

We average the rollout entries from the first $h$ action-token rows to the
256 base-image patch columns and reshape them into a $16\times16$
saliency map. Register-token columns are excluded because registers encode
global embodied context but do not retain a fixed image-grid
correspondence. The saliency map is bilinearly upsampled to the original
image resolution before window search.

For multiple camera views, rollout maps are computed independently for
each view. We select the single view--window pair with the highest
localization score, so an uncertain replanning step introduces at most one
additional crop encoding.

\subsection{Contrastive Window Search}

We search square candidate windows with side lengths
\begin{equation}
    \mathcal{S}
    =
    \{84,112,140\}\ \text{pixels}
\end{equation}
on the $224\times224$ input image. Windows are evaluated using a
14-pixel stride, aligned with the SigLIP patch size.

For candidate window $\mathbf{b}$, let $\rho(\mathbf{b})$ denote the
context window obtained by expanding its side length by a factor of
$1.25$. We select
\begin{equation}
    \mathbf{b}^{*}
    =
    \arg\max_{\mathbf{b}\in\mathcal{B}}
    \left[
        \frac{1}{|\mathbf{b}|}
        \sum_{u\in\mathbf{b}}S_t(u)
        -
        \frac{1}{|\rho(\mathbf{b})\setminus\mathbf{b}|}
        \sum_{u\in\rho(\mathbf{b})\setminus\mathbf{b}}
        S_t(u)
    \right].
    \label{eq:crop-window-app}
\end{equation}
The contrastive score favors regions whose internal attention is
distinctive relative to their local surroundings, discouraging both
diffuse full-image crops and isolated attention peaks.

The selected box is padded by 10\% on each side, clipped to the image
boundary, cropped from the original observation, and resized to
$224\times224$. Its normalized coordinates are passed through a
two-layer crop-position MLP,
\begin{equation}
    4
    \rightarrow
    256
    \rightarrow
    d_{\mathrm{LLM}},
\end{equation}
with a GELU activation. The resulting position embedding is added to every
projected crop patch token.

\subsection{Invalid Rollout Handling}

A rollout is considered invalid if any of the following conditions holds:
\begin{itemize}
    \item the saliency map contains non-finite values;
    \item its total attention mass is smaller than $\epsilon=10^{-6}$;
    \item window search produces no finite candidate score;
    \item the selected coordinates are degenerate or lie outside the image
          after clipping; or
    \item the final crop has a side length smaller than 56 pixels.
\end{itemize}

For an invalid rollout, crop refinement is skipped and the policy executes
the action prediction obtained from the base-view path. The same rule is
used during joint training and deployment, preventing ground-truth
fallbacks from introducing a train--test discrepancy.

\section{Key Implementation Configuration}
\label{app:key-config}

Key training and inference configurations are listed in Table.~\ref{tab:app-config}.

\begin{table}[t]
    \centering
    \small
    \setlength{\tabcolsep}{5pt}
    \begin{tabular}{ll}
        \toprule
        Configuration & Value \\
        \midrule
        LIBERO training data
            & 40 tasks, 2,000 trajectories \\
        Register tokens
            & 4 \\
        Input resolution
            & $224\times224$ \\
        Register adaptation
            & 20K steps: 4K register-only/16K SigLIP \\
        Ground-truth crop alignment
            & 10K steps \\
        Joint refinement
            & 40K steps \\
        LoRA rank
            & 32 \\
        Action samples for gating
            & $K=4$ \\
        Flow integration
            & 10-step Euler \\
        Rollout denoising steps
            & $\{3,6,9\}$ \\
        Rollout layers
            & All 18 joint-attention layers \\
        Candidate crop sizes
            & $\{84,112,140\}$ pixels \\
        Window-search stride
            & 14 pixels \\
        Context expansion
            & $1.25\times$ \\
        Final crop padding
            & 10\% \\
        Target trigger rate
            & Approximately 30\% per embodiment \\
        \bottomrule
    \end{tabular}
    \caption{Training and inference configuration for AtVLA.}
    \label{tab:app-config}
\end{table}

\section{More Demonstrations}
\begin{figure*}
    \centering
    \includegraphics[width=1\linewidth]{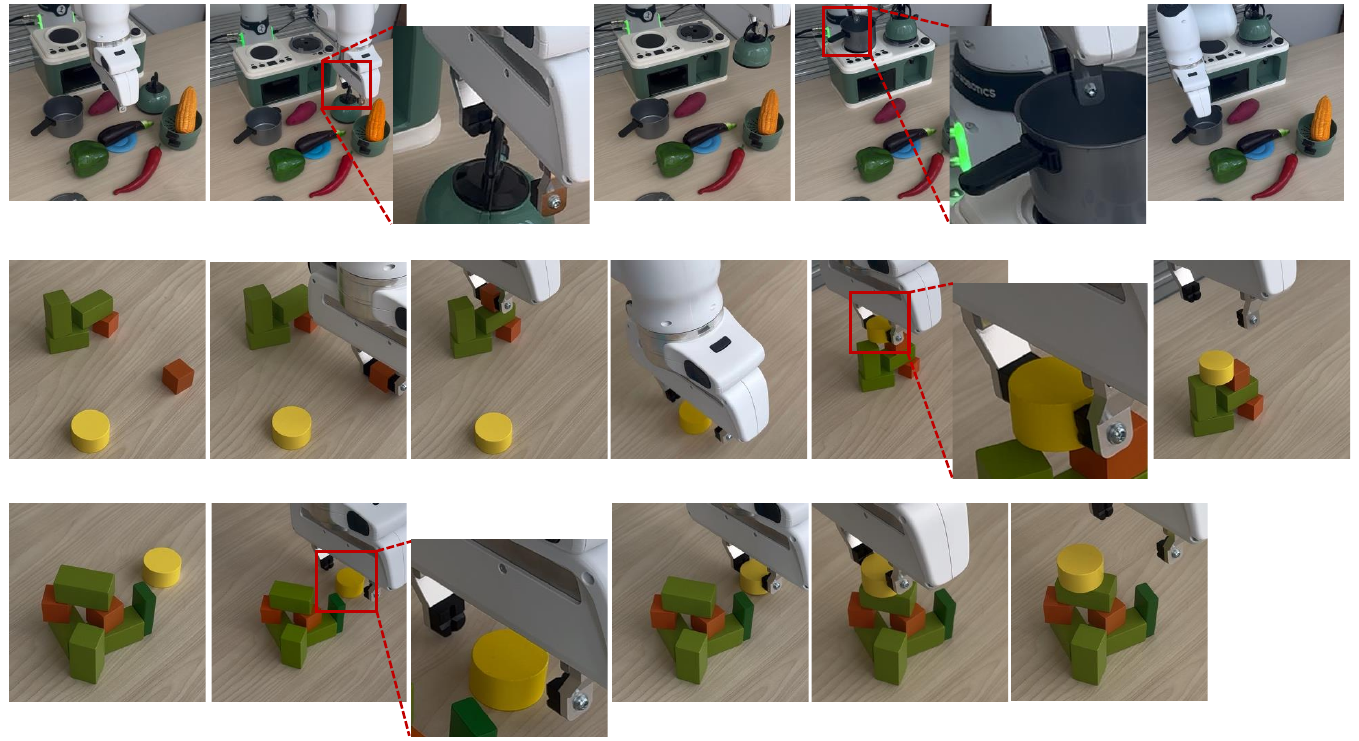}
    \caption{Additional real-world demonstrations of AtVLA. Each row presents a temporally ordered manipulation rollout. }
    \label{fig:more-demonstrations}
\end{figure*}

Figure~\ref{fig:more-demonstrations} provides additional qualitative
rollouts illustrating how AtVLA combines global scene understanding with
selective high-resolution perception. The full observation remains
available throughout execution, allowing the policy to preserve the task
instruction, workspace layout, and relationships among multiple objects.
When the action prediction becomes uncertain near a grasp or placement
event, the refinement branch concentrates additional visual capacity on
the corresponding interaction region rather than repeatedly processing
irrelevant background details.

\end{document}